\documentclass[10pt,journal,compsoc]{IEEEtran}
\usepackage{times}
\usepackage{epsfig}
\usepackage{graphicx}
\usepackage{amsmath}
\usepackage{amssymb}
\usepackage[ruled]{algorithm2e}
\usepackage{algpseudocode}
\usepackage{caption}
\usepackage{subfigure}
\usepackage{booktabs}
\usepackage{tabularx}
\usepackage{xcolor}
\usepackage{multirow}
\usepackage{cite}
\usepackage{hyperref}

\begin{document}
%

\title{AdaPose: Towards Cross-Site Device-Free Human Pose Estimation with Commodity WiFi}

%
%
%

\author{Yunjiao Zhou,
        Jianfei Yang*,
        He Huang,
        Lihua Xie,~\IEEEmembership{Fellow,~IEEE}
\thanks{Y. Zhou, J. Yang, H. Huang, and L. Xie are with the School of Electrical and Electronics Engineering, Nanyang Technological University, Singapore (yunjiao001@e.ntu.edu.sg; yang0478@e.ntu.edu.sg; he008@e.ntu.edu.sg; 
elhxie@ntu.edu.sg; ).}
\thanks{*J. Yang is the corresponding author.}  }

%
%

\markboth{Preprint}%
{Zhou \MakeLowercase{\textit{et al.}}: AdaPose: Towards Cross-Site Device-Free Human Pose Estimation with Commodity WiFi}
%



\maketitle

\begin{abstract}
WiFi-based pose estimation is a technology with great potential for the development of smart homes and metaverse avatar generation. However, current WiFi-based pose estimation methods are predominantly evaluated under controlled laboratory conditions with sophisticated vision models to acquire accurately labeled data. Furthermore, WiFi CSI is highly sensitive to environmental variables, and direct application of a pre-trained model to a new environment may yield suboptimal results due to domain shift. In this paper, we proposes a domain adaptation algorithm, AdaPose, designed specifically for weakly-supervised WiFi-based pose estimation. The proposed method aims to identify consistent human poses that are highly resistant to environmental dynamics. To achieve this goal, we introduce a Mapping Consistency Loss that aligns the domain discrepancy of source and target domains based on inner consistency between input and output at the mapping level. We conduct extensive experiments on domain adaptation in two different scenes using our self-collected pose estimation dataset containing WiFi CSI frames. The results demonstrate the effectiveness and robustness of AdaPose in eliminating domain shift, thereby facilitating the widespread application of WiFi-based pose estimation in smart cities.

\end{abstract}


%
\IEEEpeerreviewmaketitle

\section{Introduction}

\begin{figure}
\centering

\includegraphics[width=1\linewidth]{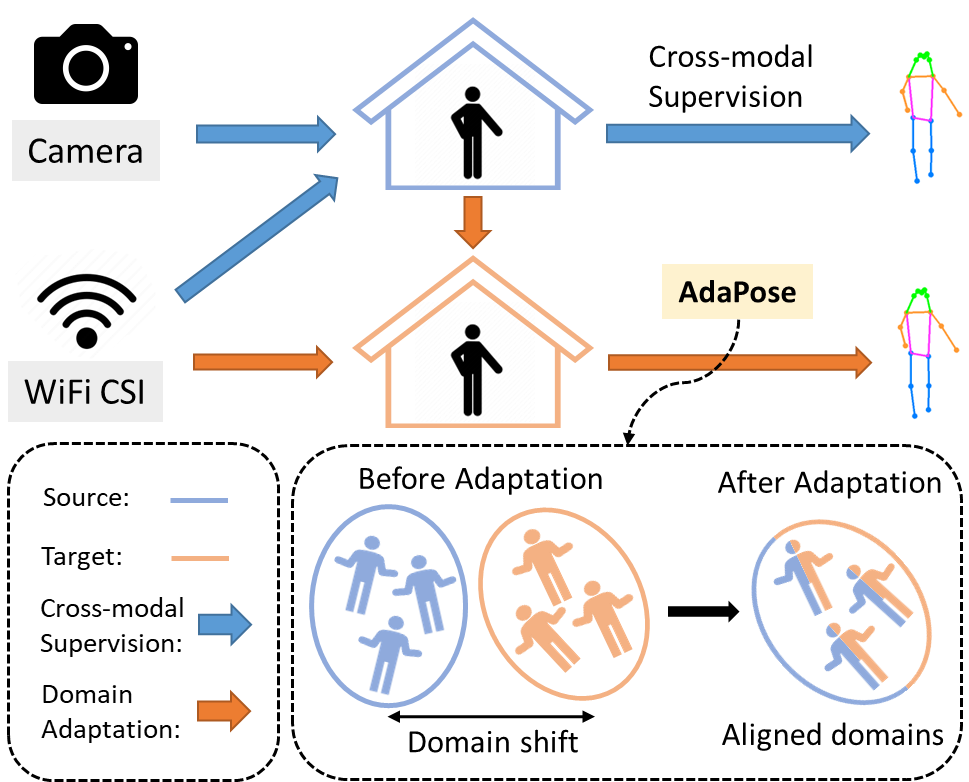}
\caption{Illustration of using domain adaption methods to improve the transferability of WiFi-based pose estimation in different environments. The bottom block shows the effectiveness of domain adaptation to align the domain shifts between the source and target domain, which transfers the knowledge learned from the source domain to improve the model's ability in a new environment.
}
\setlength{\belowdisplayskip}{0pt}
\label{fig:fig1}
\end{figure}
Human pose estimation has become a critical technology in a wide range of applications, including motion analysis, virtual reality, and healthcare~\cite{niu2021understanding}. However, the primary approaches for pose estimation, which rely on computer vision techniques\cite{andriluka20142d,toshev2014deeppose} or wearable devices\cite{mcgrath2020body, bianchi2010barometric, schiboni2021context}, face practical challenges that limit their wide applications in certain scenarios. Vision-based solutions raise severe privacy concerns due to the collection of fine-grained appearance information, while wearable devices are often inconvenient and restricted to specific use cases. Therefore, there is a growing need for non-invasive and user-friendly solutions for pose estimation. Recently, WiFi-based methods have emerged as a promising alternative that is cost-effective, privacy-preserving, and device-free. These methods exhibit exceptional and competitive performance for the widespread of human pose estimation.


Current WiFi-based pose estimation methods\cite{wang2019can,jiang2020towards, ren2021winect,yang2022metafi,zhou2023metafi++}, primarily employ deep learning techniques to extract features from WiFi CSI and predict human joint positions via regression, demonstrating promising results in controlled laboratory environments with ample labeled data. However, when these models are deployed in more private settings, such as smart homes and hospitals without access to labeled data, their performance can degrade significantly due to the high dependence of WiFi CSI on environmental dynamics, as depicted in Fig.~\ref{fig:fig1}. To address this issue, researchers have incorporated domain adaptation algorithms into WiFi-based sensing tasks, including human activity recognition\cite{zou2018robust,wang2019wicar,jiang2018towards}, gesture recognition\cite{zou2018joint,zhang2021unsupervised, wang2022airfi}, and human localization\cite{chen2020fido,chen2022fidora}. These algorithms aim to minimize the domain shifts between different environments, thereby improving performance in new settings. Despite the demonstrated effectiveness of domain adaptation in WiFi settings, no specific domain adaptation algorithm has been proposed for WiFi-based pose estimation, which is more intricate, requiring the regression of complex skeletons with limited and coarse data. Thus, the objective of this paper is to introduce a customized domain adaptation algorithm for WiFi-based pose estimation.

Nonetheless, in the field of WiFi-based pose esitimation, there are significant challenges in improving their transferability across different environments, hindering the real-world applications. 
 The first challenge arises due to the intricate nature of pose estimation, requiring models to predict the 2D positions of 17 complex human joints through regression. This necessitates models to extract subtle yet representative features from the limited and coarse WiFi CSI data, which pales in comparison to the fine-grained image data with three channels of color representation. This challenge is particularly challenging to overcome due to the inherent limitation of WiFi data to capture subtle joint positions and movements accurately.
Another major challenge stems from the substantial presence of environmental interference factors in WiFi data, which impairs the generalization ability of pose estimation models in new environments. These environmental factors exhibit a much stronger presence in WiFi data compared to dynamic human motion, thereby dominating the data and leading to the potential alignment of environmental information instead of useful human dynamic information.
Therefore, resisting the environmental dependence of WiFi CSI for the fine-grained pose estimation task is hard but crucial to enhance the transferability of pose estimation models across varying environments, where no existing work has adequately addressed this issue.

To overcome the challenges of WiFi-based pose estimation, traditional model-based or signal processing methods are inadequate due to their inability to effectively model complex human motion information and extract useful features. In contrast, deep learning approaches offer an end-to-end solution that can extract fine-grained features and capture human motion information present in the limited and coarse WiFi CSI data. 
As for the second challenge of transferability, domain adaptation regression (DAR) is highly sensitive to feature scale, as it lacks the softmax function to normalize the activation values of different categories. Consequently, existing WiFi domain adaptation methods including both adversarial-based and discrepancy-based approaches, tending to align domain shifts at the feature level,  can result in false alignment caused by scale changes, which does not align with true human motion. Therefore, there is a need to explore domain adaptation methods at the mapping level that are more suitable for regression tasks, and align the mapping relationship between the two domains while ensuring the scale remains unchanged to narrow the gap between them and mitigate the domain shift problem in fine-grained pose estimation tasks.

Therefore, we propose a novel WiFi-enabled adaptive pose estimation framework, named AdaPose, which extracts the domain-invariant relationships at the mapping level to predict accurate human poses consistent to environmental dynamics, without a large scale of annotation for a new environment. Commercial off-the-shelf (COTS) WiFi devices are served as transmitter and receivers to capture WiFi CSI data that contains human motion information. In the source domain, we take help from the advanced vision modal, HRNet~\cite{sun2019deep}, to generate 2D skeletons, serving as the ground truth of WiFi modal with cross-modal supervision. On the other hand, a Consistency Loss is designed to represent the modality mapping from WiFi CSI to human skeletons. It explores the complex relationships between the input and output of the source and target domains to find the invariant mapping rules that are less affected by domain shift. By designing ratio relationships rather than absolute feature intensity information, the feature scale can remain stable during the training, making sure that the spatial range of the target can be enlarged and mapped to the entire feature space. Experiments demonstrate the effectiveness and consistency of AdaPose for domain adaption in two scenes on our self-collected pose estimation dataset. 

Our contributions can be summarized as follows,
\begin{itemize}
    \item We propose AdaPose to identify consistent human poses with high robustness to environmental dynamics. To the best of our knowledge, AdaPose is the first solution to address cross-domain WiFi-based human pose estimation.
    \item We innovatively design a Mapping Consistency Loss to align the features of the source and target domain based on their inner consistency between input and output, suitable to deal with regression tasks.
    \item Extensive real-world experiments show the robust domain adaptation ability of AdaPose on unsupervised and semi-supervised learning.
\end{itemize}
\section{RELATED WORK}
\subsection{WiFi CSI Sensing}
WiFi Channel State Information (CSI)~\cite{ma2019wifi} is a valuable resource that captures the changes in amplitude and phase of wireless signals as they interact with objects and human beings in the surrounding environment. By analyzing these subtle changes, WiFi CSI offers fine-grained information that can be harnessed as a promising modality for indoor sensing~\cite{yang2023sensefi}. In contrast to traditional WiFi measurements that primarily focus on signal strength or signal quality, CSI provides a deeper level of understanding about the wireless propagation environment. To extract WiFi CSI, WiFi devices equipped with multiple antennas are employed to exploit the multi-path propagation, where wireless signals bounce off objects and surfaces, causing variations in the amplitude and phase of the received signal. The CSI measurement involves obtaining the complex-valued channel response, which consists of amplitude and phase information at each subcarrier of the WiFi signal.

Once the CSI is obtained, it can be processed and analyzed to extract valuable insights about the environment and the objects within it. With its low-cost, ubiquitous and privacy-preserving nature, WiFi CSI has attracted considerable attention in various sensing applications, such as activity recognition~\cite{wang2015understanding, wang2017device, wang2018spatial,zou2018deepsense,yang2018carefi,zou2019wificv,ji2022sifall}, localization~\cite{wang2018wifi,zhou2018device}, gesture recognition~\cite{yang2019learning,zou2018gesture,yang2022robustsense,yang2022autofi}, human identification~\cite{zou2018identification,wang2022caution,deng2022gaitfi}, and people counting~\cite{zou2018device,FreeCount}, and pose estimation~\cite{wang2019can,wang2019person, yang2022metafi}. For example, CARM~\cite{wang2015understanding} proposes two theoretical models to quantitatively correlate CSI dynamics with human activities, accurately predicting human activities with performance greater than 96\%. While Zhou et al.~\cite{zhou2018device} leverage deep neural networks to learn CSI fingerprint patterns, enabling precise object locations with low mean distance errors. These studies highlight the potential of WiFi CSI for activity recognition and localization tasks.

Unlike other perceptual tasks, pose estimation poses unique challenges due to its need to predict multiple joint positions with limited information. Existing methods in this field can be categorized into 2D and 3D human skeleton estimation. 2D pose estimation typically takes use of images and mature algorithms to generate the landmarks, making it relatively easy and commercially feasible to implement. For instance, WiSPPN~\cite{wang2019can} designs a convolutional neural network that captures the pose adjacency matrix while preserving the generalization ability of 2D poses. WiFi CSI has shown accurate 2D pose estimation performance in static environments. On the other hand, precise 3D ground truth is challenging to obtain automatically due to the requirement for depth information. To extract accurate 3D poses, many works seek help from various devices, such as Microsoft Kinect 2~\cite{ren2021winect} and VICON Camera~\cite{jiang2020towards}, to guide the 3D pose estimation of WiFi-based model. Nonetheless, achieving accurate 3D pose estimation still poses difficulties, particularly in attaining high-dimensional resolution. Despite the significant progress in WiFi-based pose estimation, there still remains considerable room for improvement when compared to the level of accuracy achieved by visual-based approaches.


\subsection{Domain Adaptation Methods}
Domain adaptation aims to develop a model from the source domain that can be effectively applied to a related target domain with dissimilar data distribution. 
This problem often arises in machine learning applications where labeled data in the target domain is scarce or expensive to obtain, while there exists an abundance of labeled data in a related source domain. In such scenarios, domain adaptation techniques play a crucial role in leveraging the knowledge from the source domain to improve the performance of models on the target domain.
A common strategy for domain adaptation involves aligning the source and target domains by minimizing the domain shift between them. One popular method is Domain Adversarial Neural Networks (DANN)\cite{ganin2015unsupervised}, which designs a domain classifier to distinguish between source and target domain samples. The network is trained to minimize both the classification loss and the domain adversarial loss simultaneously through a gradient reversal layer. Discrepancy-based method is another popular approach, like Maximum Mean Discrepancy (MMD)\cite{tzeng2014deep}, which minimizes the maximum mean distance between the source and target distributions to gather their feature spaces.

While existing domain adaptation methods have been primarily proposed and extensively studied in the computer vision field, where informative images with fine-grained appearance and color representations are available, applying them to WiFi CSI sensing poses unique challenges. The simple data structure and limited information content of WiFi CSI measurements can lead to overfitting and misalignment problems when directly applying traditional domain adaptation methods. As such, researchers have started exploring the robustness of WiFi CSI to environmental changes in activity recognition~\cite{zou2018robust, wang2019wicar, li2023caring,zou2018robust,yang2020mobileda} and localization~\cite{chen2020fido, chen2022fidora, li2021dafi}. For example, WiADG~\cite{zou2018robust} adapted an adversarial-based domain adaptation algorithm into a human gesture recognition model, achieving accurate and consistent results under environmental dynamics. Similarly, ~\cite{chen2020fido} designed a domain-adaptive classification framework for human localization that used a Variational Autoencoder (VAE) to augment data and joint classification-reconstruction structure to improve consistency between different users. These studies demonstrate the potential of domain adaptation techniques in improving the robustness of WiFi-based activity recognition and localization tasks. However, it is worth noting that these approaches mainly focus on classification tasks and neglect to conduct in-depth research on the specific domain adaptation characteristics of WiFi CSI.

Therefore, it is necessary to propose a customized domain adaptation algorithm for WiFi-based pose estimation. By exploring the domain adaptation characteristics of WiFi CSI sensing, it is possible to enhance the performance and generalization capabilities of pose estimation models, narrowing the gap between WiFi-based approaches and their visual-based counterparts.

\section{METHOD}
\subsection{CSI Sensing System Design}
To facilitate the widespread implementation of WiFi sensing systems in smart home, it is crucial to develop solutions that are convenient and cost-effective on a large scale. While existing CSI-based systems adopt either laptops with Intel 5300 NIC cards ~\cite{ren2021winect, wang2015understanding}, or commercial WiFi devices with limited subcarriers to capture the raw CSI data~\cite{wang2019can,wang2021point}, they lack for the level of scalability and cost-effectiveness. In this study, we develop a CSI sensing system by employing a camera and two COTS WiFi routers, specifically TP-Link N750, which serve as one transmitter and three receivers, equipped with 3 pairs of antennas. The configuration allows for the capture of up to 114 subcarriers of CSI for each antenna pair, thereby providing a comprehensive and accurate dataset for analysis. The system is built on the openwrt platform operating system, enabling the synchronization of CSI and image signals for cross-modal supervision. 

Specifically, CSI is a crucial parameter in wireless communication that describes the channel properties between the pairs of transmitter and receiver. It encapsulates the amplitude attenuation and phase shift of WiFi signals across subcarriers, which in turn reflects human motion and the environmental layout along the transmission path. Based on orthogonal frequency division multiplexing (OFDM) theory~\cite{stuber2004broadband}, CSI data can be collected simultaneously from multiple antennas and multiple subcarriers, offering a comprehensive view of indoor scenes. Consequently, our CSI frame at each time instance consists of $3 \times 114$ amplitude and phase measurements, providing fine-grained information compared to the Received Signal Strength (RSS). Mathematically, the CSI measurement of each subcarrier can be expressed using the Channel Impluse Response (CIR) $H(t)$:
\begin{equation}
H(t)=\sum_{n=1}^{N_p}a_n(t) e^{j2\pi f_c \tau_n(t)},
\end{equation}
where $N_p$ is the number of paths reaching to the receiver, $f_c$ denotes the carrier frequency, $a_n(t)$ and $\tau_n(t)$ represent the complex attenuation and time delay of the $n$-th path at time $t$, respectively. Due the sensitivity of phase to small variations and noise, we only leverage amplitude as CSI data, which is more robust to environmental dynamics. According to the sampling rate of WiFi devices and camera, one image is corresponding to 32 frames of CSI, denoted as $\mathbf{x} \in \mathbb{R}^{3\times114\times32}$.

After obtaining the synchronized video and CSI frames, we adopt AI model to realize cross-modal pose estimation task. To begin with, HRNet~\cite{sun2019deep} processes the images to generate 2D human skeletons with 17 joints, denoted as $\mathbf{y} \in \mathbb{R}^{17\times2}$. Trained with large COCO dataset~\cite{lin2014microsoft}, the results are accurate and robust enough to serve as the ground truth for WiFi-based model. As for WiFi CSI, we leverage our previous network WPNet~\cite{yang2022metafi} as the baseline, which uses convolutional layers and residual blocks~\cite{he2016deep} to extract the CSI features $\mathbf{f} \in \mathbb{R}^{512\times17\times17} $ with human motion information. Then a bottleneck and average pooling layer regresses for the final predictions of human poses $\hat{\mathbf{y}}\in \mathbb{R}^{17\times2}$. During the training process, Mean Squared Error (MSE) loss is used as the learning objective to optimize the model parameters: 
\begin{equation}
    \mathcal{L}_{MSE}=||\hat{\mathbf{y}}-\mathbf{y}||^2_2.
\end{equation}

In summary, we have developed a low-cost and scalable WiFi-based sensing system enabling the simultaneous collection of multiple subcarriers and antennas. By integrating cross-modal supervision and AI techniques, we have demonstrated the potential of our system for fine-grained human pose estimation, making it an attractive solution for a wide range of smart home applications. 

\subsection{AdaPose: WiFi-enabled domain adaptation framework}
Since WiFi CSI data is highly susceptible to the environmental noise and changes, the performance of the CSI sensing system can significantly deteriorate when transitioning to new scenes. In addition, it is inpracticle to collect the large amount of labled data whenever the system is deployed in an untrained environment. To overcome these issues, we propose a customized domain adaptation framework, called AdaPose, to improve the robustness and consistency of the system to various settings. A novel Mapping Consistency Loss is designed to consider the relative relationships of output and input from the source and target domains at the mapping level without changing the feature scale, narrowing the domain shift through the inherent connection between the two domains. 
AdaPose is a general adaptation method for WiFi-based pose estimation, enabling the CSI sensing systems with strong ability of domain alignments for a new environment when only a few labels are available in target domain.

\begin{figure*}
\centering

\includegraphics[width=1\linewidth]{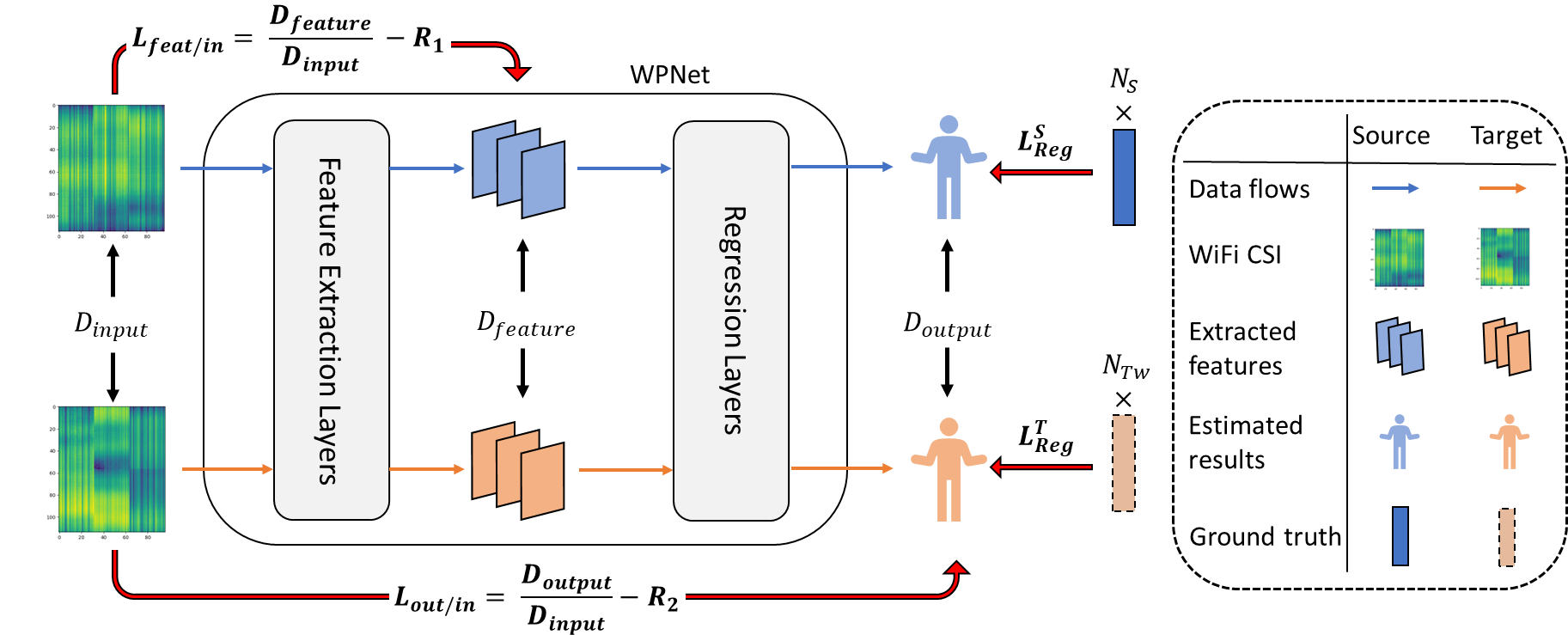}
\caption{The main structure of AdaPose.}
\setlength{\belowdisplayskip}{0pt}
\label{fig:fig2}
\end{figure*}

The approach assumes access to a labeled source domain, denoted as $\mathcal{D_S}={(\mathbf{x}^S_{i}, \mathbf{y}^S_{i})}_{i=1}^{N_S}$, which contains $N_S$ frames of WiFi CSI data and corresponding human skeleton ground truth data. Additionally, a labeled target domain, denoted as $\mathcal{\widetilde{D}_{T}}={(\mathbf{\widetilde{x}}^T_{j}, \mathbf{\widetilde{y}}^T_{j})}_{j=1}^{\widetilde{N}_T}$, is provided with $\widetilde{N}_T$ frames of CSI and ground truth data, while an unlabeled target domain, denoted as $\mathcal{D_{T}}={(\mathbf{x}^T_{k})}_{k=1}^{N_T}$, contains only $N_T$ frames of CSI data. While both $N_S$ and $N_T$ are assumed to be sufficiently large for the training process, $\widetilde{N}_T << N_S$ indicates that only a small amount of labeled data is available in the target domain. The core idea behind weakly-supervised domain adaptation is to leverage this limited labeled data in the target domain, along with the abundant labeled data in the source domain, to learn a domain-invariant representation that can generalize well to the target domain. This approach aims to address the challenge of poor performance when transferring CSI-based pose estimation models between different environments, by enabling the model to adapt to the target environment using limited labeled data.

To start with, the framework of AdaPose is illustrated in Fig.~\ref{fig:fig2}, where the source and target data share the same structure of the feature extraction layers and regression layers. 
The feature extraction layers $E(\cdot)$ consist of several convolutional layers and residual blocks to extract the useful information of CSI, while the regression fuction $R(\cdot)$ is realized by the bottleneck layers with the average pooling. 
Specifically, the CSI data $x^S_i,\widetilde{x}^T_j$ and $x^T_k$ are input into the feature extractor to generate the extracted features $f^S_i,\widetilde{f}^T_j$ and $f^T_k$, and then are processed by the regression layers to obtain the estimated human pose results $\hat{y}^S_i,\hat{\widetilde{y}}^T_j$ and $\hat{y}^T_k$. To guide the mapping of the network, we develop the regression loss by optimizing the error between the predicted results and the ground truth in source and target labeled data, respectively: 
\begin{equation}
\begin{split}
    \mathcal{L}_{Reg}^S(E,R)=\frac{1}{N_S}\sum_{i=1}^{N_S}||\hat{y}^S_i-\mathbf{y}^S_{i}||^2_2,
    \\
    \mathcal{L}_{Reg}^{\widetilde{T}}(E,R)=\frac{1}{\widetilde{N}_T}\sum_{j=1}^{\widetilde{N}_T}||\hat{\widetilde{y}}^T_j-\mathbf{\widetilde{y}}^T_{j}||^2_2.
\end{split}
\end{equation}
The objective of regression loss can be summarized as the following optimization:
\begin{equation}
     \mathcal{L}_{Reg}(E,R)= \mathcal{L}_{Reg}^S(E,R)+\mathcal{L}_{Reg}^{\widetilde{T}}(E,R).
\end{equation}
Through the supervised learning process, the model is trained to generate accurate human poses via sufficient source data, while the target labeled data is used to fine-tune the model on the target domain, so that the learned representation can be adjusted to better fit the target data distribution.

\subsection{Mapping Consistency loss} After obtaining the well-trained model on source domain, how to mitigate the domain shift to improve the performance on the target domain remains an essential issue. Based on the above finding that feature scales vary a lot in the regression task, it can not be a useful strategy for the existing methods to align domain shift at the feature level, which may cause the scale misalignment. Thus it is necessary for WiFi-based pose estimation to find a new adaptation approach to align the domain shift at the non-feature level.

In fact, though source and target have different data distribution, the mapping rule of the pose estimation remains the same, that is to transfer the modality of data from CSI to skeleton positions. So there exists the consistent relationship between the source and target domain. Regardless of the discrepancy of the two domains, the network mapping is to transfer the input to the output data based on certain consistent patterns~\cite{taghiyarrenani2023multi}. Therefore, we aim to utilize the consistency between the input and output of the source and target domain to align the mapping stage of the model, so as to minimize the domain shift without changing the feature scales for regression task. To make sure the input relationship between the source and target domains can be retained and reflected in the output after mapping, we design a ratio of output and input distribution discrepancy between the source and target domains to represent the consistency of the model.

To find the consistency more precisely, we break the network mapping into two parts, the mapping between the extracted feature and input, and the mapping between the output and input, so as to align the feature extractor layers and regression layers respectively.
Mathematically, we adopt the Maximum Mean Discrepancy (MMD), which is often used to measure the distance between source and target distributions, to calculate domain distribution discrepancy of input, extracted feature and output, denoted as $D_{input}, D_{feature}$ and $D_{output}$. The formulations can be expressed as:
\begin{equation}
\begin{split}
    D_{input}=||\frac{1}{N_S}\sum_{i=1}^{N_S}\phi(x^S_i)-\frac{1}{N_T}\sum_{k=1}^{N_T}\phi(x^T_k)||^2_H,
    \\
     D_{feature}=||\frac{1}{N_S}\sum_{i=1}^{N_S}\phi(f^S_i)-\frac{1}{N_T}\sum_{k=1}^{N_T}\phi(f^T_k)||^2_H,
     \\
      D_{input}=||\frac{1}{N_S}\sum_{i=1}^{N_S}\phi(\hat{y}^S_i)-\frac{1}{N_T}\sum_{k=1}^{N_T}\phi(\hat{y}^T_k)||^2_H,
\end{split}
\end{equation}
where $\phi(\cdot)$ refers to a kernel function that maps data points from the input space to a reproducing kernel Hilbert space (RKHS). Then we calculate the ratio between the two distances to represent the consistency of the mappings rules, formulated as:
\begin{equation}
\begin{split}
    \mathcal{R}_{feat/in}=\frac{D_{feature}}{D_{input}},
    \\
     \mathcal{R}_{out/in}=\frac{D_{output}}{D_{input}}.
\end{split}
\end{equation}
Based on the consistency relationships of the mapping, the ratios of two distribution discrepancies are supposed to be stable and robust to the scale changes. AFN~\cite{xu2019larger} introduces a stepwise constraint to regulate the norm value in a progressive manner, but it may amplify the effect of each sample, including the outliers, making the training unstable. Thus, we compare the raitos with restrictive scalars $r_1,r_2$ to assess the consistency losses at the mappling level:
\begin{equation}
\begin{split}
    \mathcal{L}_{feat/in}(E)=L_d(\mathcal{R}_{feat/in},r_1),
    \\
     \mathcal{L}_{out/in}(E,R)=L_d(\mathcal{R}_{out/in},r_2),
\end{split}
\end{equation}
where $L_d(\cdot,\cdot)$ denotes the $L_2$-distance, and the hyperparameters $r_1,r_2$ ensure that the consistency ratios of two features from different stages converge towards a shared equilibrium, which are obtained through experimentation. The mapping consistency losses can gradually make the ratios approach a fixed value, so that the relationships in the input of the mapping are preserved in the output, thus aligning the shift of the mappings between the source and target domains.

Finally, the goal of AdaPose is to learn the domain-invariant representations through minimizing both regression and consistency losses. The optimization function of AdaPose is defined as:
\begin{equation}
    \underset{E,R}{\min} \; \mathcal{L}_{Reg}(E,R)+\alpha\mathcal{L}_{feat/in}(E)+\beta\mathcal{L}_{out/in}(E,R),
\end{equation}
where $\alpha$ and $\beta$ are the hyper-parameters to achieve the trade-off of the learning objectives. With a small amount of target labeled data, AdaPose is able to obtain high accuracy in the target domain, aligning the domain shifts at the mapping level.   
\section{EXPERIMENT}
\subsection{Setup}
\begin{figure}
\centering
\includegraphics[width=1\linewidth]{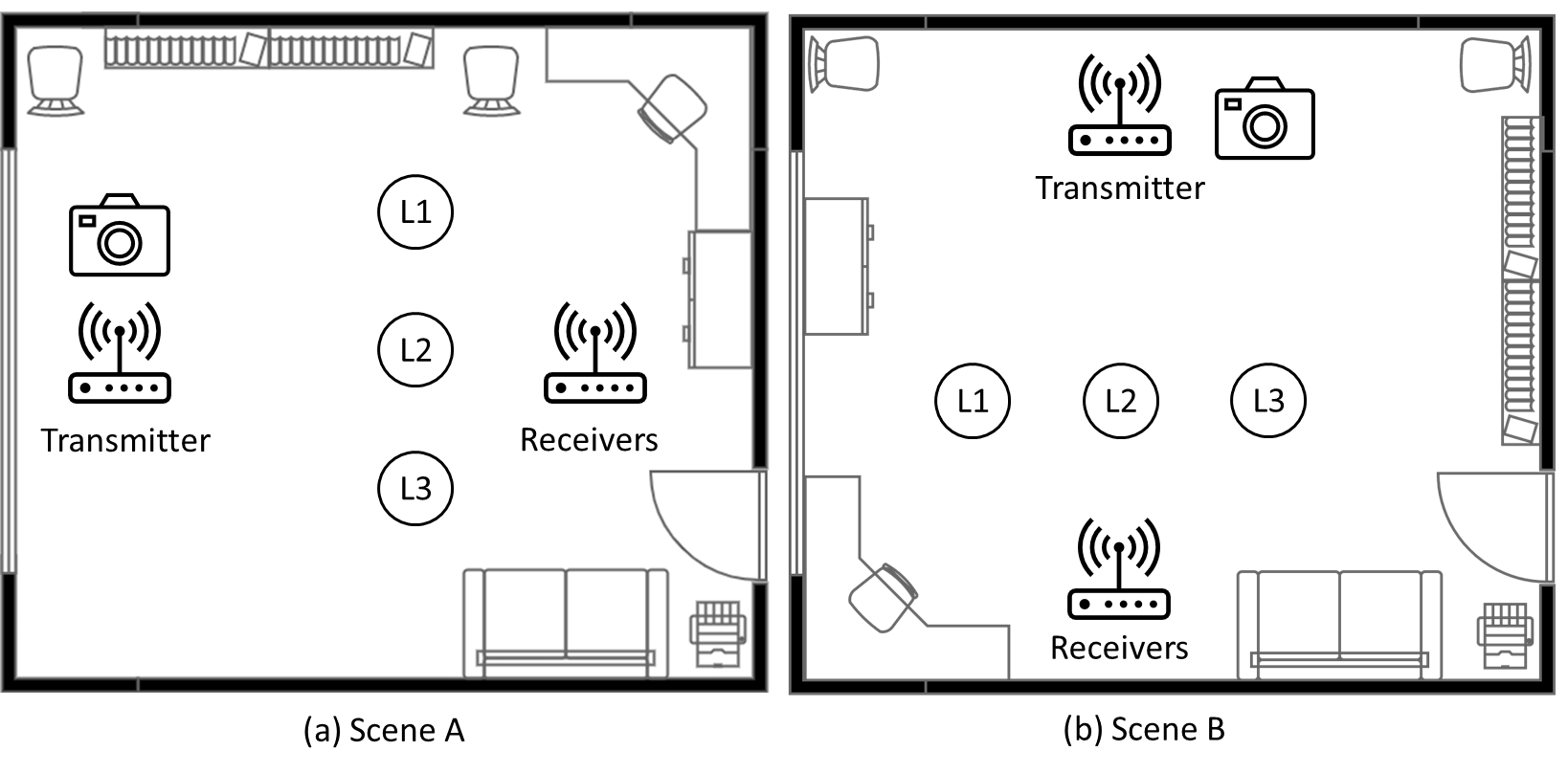}
\caption{Layout diagrams of WiFi devices in two environments.}
\setlength{\belowdisplayskip}{0pt}
\label{fig:fig3}
\end{figure}

\noindent\textbf{System setup.} In our experiment, we deployed a WiFi system consisting of two TP-Link N750 routers, placed 3.5m apart as the transmitter and receivers, respectively. The camera is placed parallel to the transmitter, collecting the vision information to generate the ground truth for the system. We set up the WiFi system in two different configurations, with the axes of WiFi devices in the two domains perpendicular to each other, contributing to scene A and scene B, as shown in Fig.~\ref{fig:fig3}. Although the data from the two domains was collected in the same area, only with the orthogonal collection directions, the transmission paths of the WiFi signals were completely different due to the changes in the layout, resulting in a large variance in CSI data between the two domains. We invite 22 volunteers to participate in the experiment and perform daily actions 2.5m from the transmitter, with 13 allocated at scene A and 9 at scene B. To increase the diversity of data, they are randomly assigned to three different locations, namely L1, L2 and L3, which simulates a more general situation and brings huge challenge for domain adaption. 

\begin{figure}
\centering
\includegraphics[width=1\linewidth]{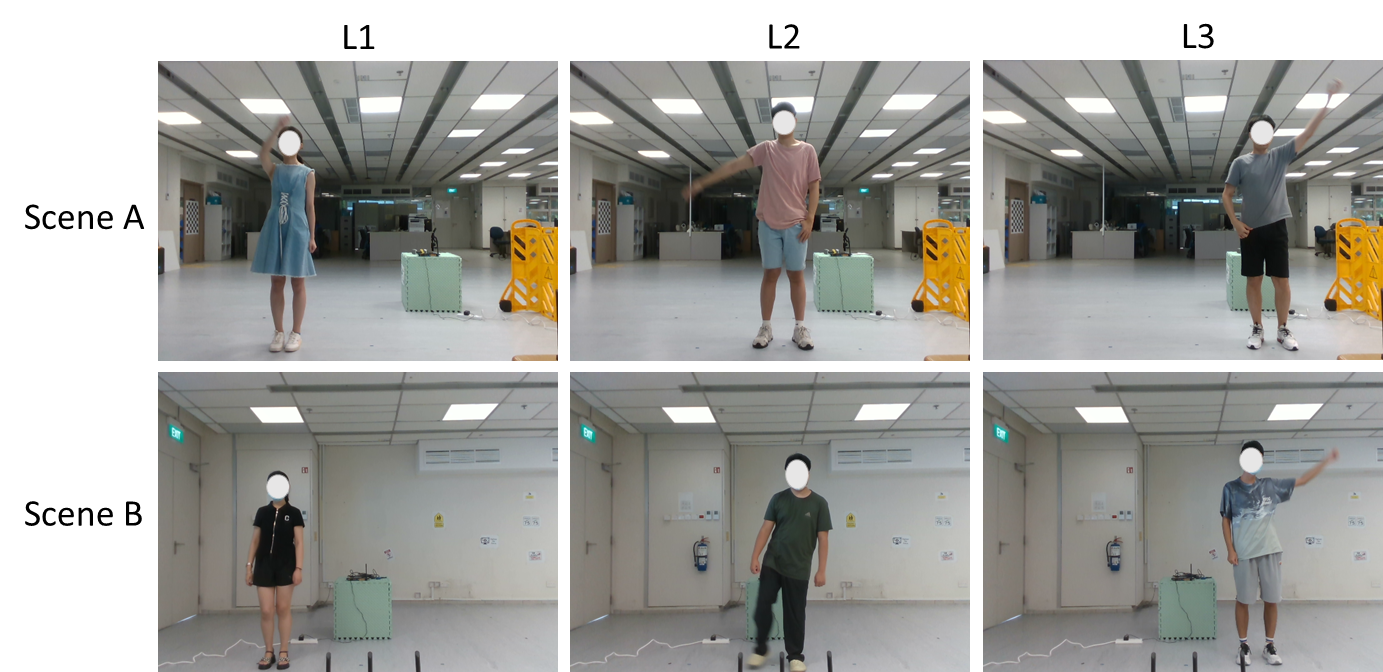}
\caption{Illustration of data collection at two domains.}
\setlength{\belowdisplayskip}{0pt}
\label{fig:fig4}
\end{figure}

During the data collection period, each volunteer is invited to perform six minutes of daily actions as illustrated in Fig.~\ref{fig:fig4}, such as waving the hands, swinging the legs, and stepping in place. Thus our dataset totally contains 132 minites of synchonized video and WiFi CSI data stream. The video and CSI signals are downsampled at rates of approximately 31 Hz and 1000 Hz, respectively, such that each video frame corresponds to 32 packets of CSI data. As a result, scene A and scene B contain 13728 and 9504 frames of video and WiFi data, respectively. Each frame of the videos collected by the camera is processed by HRNet~\cite{sun2019deep} to generate the labels of human poses for WiFi CSI data. Scene A and scene B alternate as the source and target domains for the domain adaptation experiments.

\noindent\textbf{Experimenatal details.} The AdaPose system is built on the PyTorch platform and optimized using the stochastic gradient descent with momentum (SGDM) algorithm for 50 epochs. We used a batch size of 16, learning rate of 0.001, and momentum of 0.9 to train the model parameters. To prevent overfitting and improve the model's generalization ability, we implemented a multistep learning rate decay, where the learning rate was reduced by 50\% at steps 20, 30, and 40. The training process was carried out on a machine with an NVIDIA GeForce GTX 3090 GPU. All results presented are the average of three rounds of experiments, conducted to ensure consistency and accuracy of the results.

To evaluate the performance of our proposed AdaPose system, we use the Percentage of Correct Keypoints (PCK) metric as the evaluation criterion, which is defined by:
\begin{equation}
    PCK_i@a = \frac{1}{N}\sum_{i=1}^NI(\frac{\Vert pd_i - gt_i \Vert_2^2}{\sqrt{rs^2 + lh^2}} \leq a ),
\end{equation}
where N is the number of joints taken as 17 in our experiments, and $I(\cdot)$ denotes the a function that returns a value of 1 when the input condition is true, and 0 when it is false. PCK measures the accuracy of keypoint detection by calculating the percentage of keypoints that fall within a certain distance threshold $a$ of the ground truth. We set the threshold $a$ to be 10\% of the length of torso between right shoulder ($rs$) and left hip ($lh$) in our experiments, as it has been shown to be a suitable value for human pose estimation tasks. By using multiple evaluation metrics, we can comprehensively evaluate the performance of our proposed AdaPose system and compare it with other state-of-the-art methods.

\noindent\textbf{Baseline methods.}
To verify the necessity of domain alignment strategy, we evaluate the transferability of the current WiFi-based pose estimation methods. Specifically, we reproduce two convolutional networks, namely WiSPPN~\cite{wang2019can} and WPNet~\cite{yang2022metafi}, using our dataset from one scene and validate their performance on another scene. Additionally, we also conduct fair comparison between AdaPose and several classical domain adaptation algorithms, including MMD\cite{tzeng2014deep}, a discrepancy-based method; DANN\cite{ganin2015unsupervised}, an adversarial-based method; and LIRR\cite{li2021learning}, the state-of-the-art weakly-supervised DA method to prove the effectiveness of our method.

\subsection{Performance evaluation}
\textbf{In a single environment.} To begin with, we evaluate the performance of AdaPose on a single domain, training and validating the model with 60\% and 20\% data of a scene respectively in a supervised manner. Then test the results on the rest of 20\% data, which are shown in Table~\ref{table:1}. In general, our method achieves a high accuracy with PCK@30 exceeding 90\% in both of the scenes, showing high-quality human pose estimation. Based on the evaluation results, scene B demonstrated a superior performance with a higher PCK@50 score of 98.49\% in comparison to scene A, which only achieved a score of 95.51\%. This indicates that scene B is more representative than scene A. One of the factors that could account for the difference in performance is that scene B has a clearer background with fewer environmental obstructions compared to scene A. 
\begin{table}[t]
\begin{center}
\caption{Pose estimation performance in the original environments (source domain).}
\label{table:1}
\scalebox{0.87}{
\begin{tabular}{c|ccccc}
            \toprule
            \textbf{Environment} & \textbf{PCK@50} & \textbf{PCK@40 } & \textbf{PCK@30} & \textbf{PCK@20} & \textbf{PCK@10} \\
            \midrule
            Scene A & 95.51\% & 93.73\% & 90.42\% & 84.03\% &  62.21\% \\
            Scene B & 98.49\% & 97.18\% & 96.62\% & 93.60\% & 75.41\%  \\
            
             \bottomrule
        \end{tabular}}
\end{center}
\end{table}

\noindent\textbf{Weakly-supervised domain adaptation.}
Despite the remarkable progress made by WiFi-based models in a single environment, their performance tends to suffer significantly when applied to new domains, mainly due to their reliance on specific environmental factors. Although a small portion of target labels are involved during training, Table.~\ref{table:2} shows a decrease of over 45\% in pure supervised learning without adaptation. Thus AdaPose is proposed to overcome this issue, eliminating the domain discrepancy at the mapping level with a small amount of target labeled data. To verify its effectiveness, we compare AdaPose to the non-adaptation methods and other popular domain adaptation algorithms via weakly-supervised learning in Table.~\ref{table:2}. We conducted experiments on two tasks, using 1\% of target domain labels for domain adaptation from A to B, and only 0.1\% of target domain labels for adaptation from B to A. This is possibly due to the fact that the B-to-A transfer task is relatively easier, and using 1\% of target domain labels in this case would result in a non-adaptation method achieving a very high performance, leaving little room for improvement by domain adaptation.

We observe that MMD and DANN achieve positive adaptive performance when transferring from scene A to scene B. However, when transferring from scene B to scene A, all DA methods exhibit negative effects. This observation can be attributed to the fact that existing mainstream DA methods, whether adversarial-based or discrepancy-based, tend to align the domain shift at the feature level, which can destabilize the adaptation process by altering the feature scale.
In contrast, our AdaPose method, which aligns the mapping of the model without changing the feature scales, outperforms the other methods with a large margin and achieves the highest PCK scores at all thresholds $a$ ranging from 10 to 50. Compared to the no adaptation method, our method achieves a significant PCK@50 improvement of over 27\% and 8\% in the A-to-B and B-to-A scenarios respectively, effectively transferring the source knowledge to the target domain. Theses results demonstrate the effectiveness of our proposed method for domain adaptation when only limited target domain labels are available.
\begin{table}[t]
\begin{center}
\caption{Weakly-supervised pose estimation performance comparison of domain adaptation between two scenes. (1\% target labels are available in A $\rightarrow$ B, while 0.1\% target labels are available in B $\rightarrow$ A.)}
\label{table:2}
\scalebox{0.73}{
\begin{tabular}{cc|ccccc}
            \toprule
            \textbf{Task} & \textbf{Method} & \textbf{PCK@50} & \textbf{PCK@40 } & \textbf{PCK@30} & \textbf{PCK@20} & \textbf{PCK@10} \\
            \midrule
            \multirow{6}*{\textbf{A}\textbf{$\rightarrow$}\textbf{B}} & WiSPPN\cite{wang2019can} & 44.75\% & 34.91\% & 23.34\% & 11.10\% & 2.50\%   \\
            & WPNet\cite{yang2022metafi} & 49.16\% & 42.68\% & 33.24\% & 18.03\% &  3.51\% \\
            &  WPNet + MMD\cite{tzeng2014deep}  & 57.68\% & 50.21\% & 38.75\% & 20.77\% & 4.03\%   \\ 
            & WPNet + DANN\cite{ganin2015unsupervised} & 61.91\% & 55.00\% & 44.95\% & 29.12\% & 7.89\% \\
            & WPNet + LIRR\cite{li2021learning} & 50.56\% & 43.91\% & 33.77\% & 17.53\% & 3.33\%  \\ 
            & Ours & \textbf{78.33}\% & \textbf{71.15}\% & \textbf{59.26}\% & \textbf{38.81}\% & \textbf{10.18}\%  \\ 
            \midrule
            \multirow{6}*{\textbf{B}\textbf{$\rightarrow$}\textbf{A}} & 
            WiSPPN\cite{wang2019can} & 31.28\% & 23.28\% & 14.56\% & 6.53\% & 1.87\% \\
            &WPNet\cite{yang2022metafi} & 39.00\% & 30.13\% & 20.16\% & 10.80\% &  3.03\% \\
            & WPNet + MMD\cite{tzeng2014deep}  & 33.10\% & 25.95\% & 17.35\% & 8.55\% & 2.18\%   \\ 
            & WPNet + DANN\cite{ganin2015unsupervised} & 38.95\% & 31.49\% & 23.10\% & 13.35\% & 3.92\% \\
            & WPNet + LIRR\cite{li2021learning} & 35.63\% &	29.21\% &	20.88\% &	10.39\% &	2.45\%   \\ 
            & Ours & \textbf{47.15}\% & \textbf{35.68}\% & \textbf{26.12}\% & \textbf{14.98}\% & \textbf{4.96}\%  \\
             \bottomrule
        \end{tabular}
        }
\end{center}
\end{table}

\noindent\textbf{Unsupervised domain adaptation.} We also evaluate the performance of AdaPose at an unsupervised setting in Table.~\ref{table:3}, which means no labels are accessible in the target domain. 
Without the guidance of target labels, the adaptation for a new environment is challenging, but our method shows promising results in both A$\rightarrow$B and B$\rightarrow$A scenarios, outperforming MMD and DANN by approximately 5\% PCK@50. For instance, our method achieves a PCK@50 of 36.58\% and 35.12\% in A$\rightarrow$B and B$\rightarrow$A scenarios, respectively, which are the highest values among all methods. Notably, our method outperforms all other methods across all PCK thresholds, indicating its effectiveness in unsupervised domain adaptation.

In a nutshell, AdaPose has shown positive effects in both unsupervised and weakly supervised scenarios, demonstrating the robustness and effectiveness of the model. Thus, the trade-off between annotation cost and transfer performance can be considered based on the application objectives. However, using a small number of target domain labels can significantly improve domain adaptation performance, where the benefits are likely to far exceed the cost.
\begin{table}[t]
\begin{center}
\caption{Unsupervised pose estimation performance comparison of domain adaptation between two scenes.}
\label{table:3}
\scalebox{0.73}{
\begin{tabular}{cc|ccccc}
            \toprule
            \textbf{Task} & \textbf{Method} & \textbf{PCK@50} & \textbf{PCK@40} & \textbf{PCK@30} & \textbf{PCK@20} & \textbf{PCK@10} \\
            \midrule
            \multirow{5}*{\textbf{A}\textbf{$\rightarrow$}\textbf{B}} & 
            WiSPPN\cite{wang2019can} & 14.02\% & 9.26\% & 5.19\% & 2.21\% & 0.30\% \\
            & WPNet\cite{yang2022metafi} & 25.25\% & 20.45\% & 13.38\% & 5.06\% &  0.73\% \\
            & WPNet + MMD\cite{tzeng2014deep}  & 28.21\% & 18.90\% & 11.29\% & 5.46\% & 1.17\%   \\ 
            & WPNet + DANN\cite{ganin2015unsupervised} & 30.55\% & 20.31\% & 11.39\% & 4.77\% & 0.73\% \\
            & Ours & \textbf{36.58}\% & \textbf{26.12}\% & \textbf{15.56}\% & \textbf{6.68}\% & \textbf{1.25}\%  \\
            \midrule
            \multirow{5}*{\textbf{B}\textbf{$\rightarrow$}\textbf{A}} &
            WiSPPN\cite{wang2019can} & 17.85\% & 12.63\% & 7.98\% & 3.65\% & 0.71\% \\
            & WPNet\cite{yang2022metafi} & 29.11\% & 22.37\% & \textbf{14.94}\% & \textbf{6.86}\% &  1.39\% \\
            & WPNet + MMD\cite{tzeng2014deep}  & 30.50\% & 20.79\% & 11.64\% & 4.72\% & 1.01\%   \\ 
            & WPNet + DANN\cite{ganin2015unsupervised} & 31.33\% & 21.82\% & 11.23\% & 3.73\% & 0.76\% \\
            
            & Ours & \textbf{35.12}\% & \textbf{24.53}\% & 14.29\% & 6.36\% & \textbf{1.63}\%  \\
             \bottomrule
        \end{tabular}}
\end{center}
\end{table}

\subsection{Performance Visualization}
\begin{figure}
\centering
\includegraphics[width=1\linewidth]{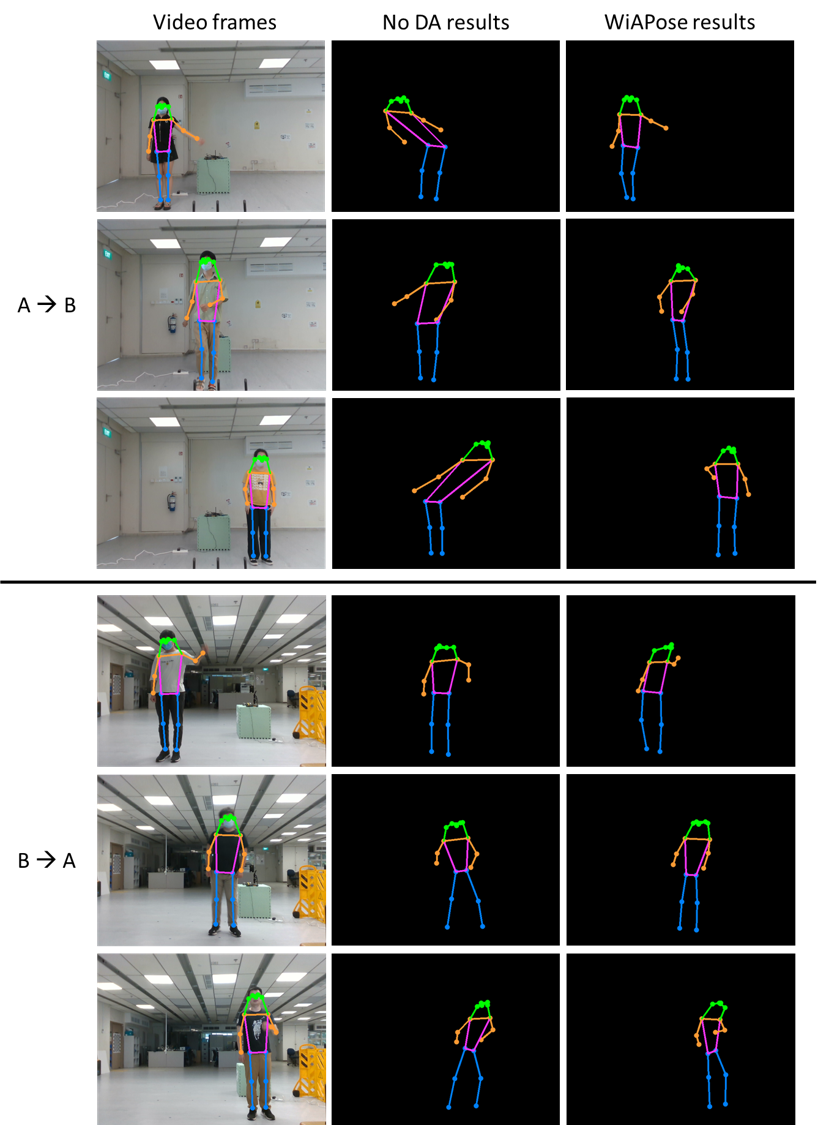}
\caption{Visualization comparision between AdaPose and no domain adaptation (DA) methods in the setting of weakly-supervised learning.}
\setlength{\belowdisplayskip}{0pt}
\label{fig:fig5}
\end{figure}

In this section, we present visualizations of our experimental results for better comparison. In Fig.~\ref{fig:fig5}, both A-to-B and B-to-A adaptation experiments of weakly-supervised setting are illustrated with no adaptation method and AdaPose, as shown in the second and third columns. While the first column demonstrates the ground truth of human poses generated by the video frames.
The results indicate that AdaPose can effectively perceive changes in the position of human subjects, whereas the no adaptation method tends to estimate the subjects mostly in the central area, due to AdaPose preserving scale information and avoiding compressing features of different scales into a standard scale during learning. Besides, the transfer from A to B demonstrates better results, with AdaPose accurately capturing subject positions and the non-DA method perceiving positional changes to some extent, despite difficulties in recognizing changes in leg position. This is due to it employing more target labels in A-to-B than B-to-A scenarios, providing stronger supervision and resulting in better performance. Though using more labels could improve domain adaptation, this comes at a higher labor and economic cost in practical applications. Therefore, AdaPose exhibits significant domain adaptation capabilities, particularly in the perception of subject positions, indicating the effectiveness of ensuring the consistency between the input and output at the mapping level.

\subsection{Ablation study}
The ablation studies are conducted to analyze the effectiveness of the proposed mapping consistency losses, including both the feature-to-input alignment loss and the output-to-input alignment loss, in the weakly-supervised and unsupervised settings. The results show that incorporating the feature-to-input alignment loss or the output-to-input alignment loss leads to significant improvements in performance, indicating the effectiveness of the mapping consistency loss. Notably, the output-to-input alignment loss shows a slightly stronger impact on the PCK scores in the unsupervised setting compared to the combined losses. However, it is practical to adopt a combination of both the feature-to-input and output-to-input alignment losses, which may be more effective in achieving robust results for wide applications. This is because each alignment loss captures different aspects of the relationship between the source and target domains. Therefore, leveraging both losses can potentially enhance the robustness of the model and enable better adaptation to new domains or tasks. 
These findings suggest that the proposed mapping consistency loss is effective in improving the transferablity of the model and aligning the source and target domains, which can be useful in various experiment settings.
\begin{table}[t]

\begin{center}
\caption{Ablation study of WiAPose on A$\rightarrow$B scenario. S-gt and T-gt mean that all the source labels and a small amount of target labels are available in the experiments, respectively. }
\label{table:5}
\scalebox{0.83}{
\begin{tabular}{cccc|cc}
            \toprule
            \textbf{S-gt} & \textbf{T-gt} & \textbf{ $\mathcal{L}_{feat/in}$} & \textbf{ $\mathcal{L}_{out/in}$} &  \textbf{PCK@50 (\%)} & \textbf{PCK@30 (\%)}\\
            \midrule
               \checkmark &  &  &  & 25.25 & 13.38 \\
              \checkmark &  & \checkmark &  & 33.16 & 12.47 \\
              \checkmark & &  & \checkmark & 35.12 & \textbf{18.94} \\
              \checkmark & & \checkmark & \checkmark &   \textbf{36.58} & 15.56\\ \midrule
             \checkmark &  \checkmark &  &  &  49.16 & 33.24 \\ 
              \checkmark & \checkmark  & \checkmark &  & 72.50  & 54.08 \\ 
              \checkmark &  \checkmark &  & \checkmark & \textbf{78.33}  & \textbf{59.26} \\
            \checkmark &  \checkmark & \checkmark & \checkmark &  76.22 & 59.22 \\ \bottomrule
        \end{tabular}}
\end{center}

\end{table}
\setlength{\tabcolsep}{1.4pt}

\section{Conclusion}
In conclusion, we proposed a WiFi-enabled pose estimation system which was demonstrated to be robust to the environmental changes. Specifically, the customized domain adaptation framework, AdaPose, is designed to align the domain shift between the source and target data with only a small amount of target labels, benefiting to the knowledge transfer to a new environment. In addition, the Consistency Loss can make sure the scale of features in the regression tasks remains stable, so as to bring great improvement of the transfer ability. The experiments show that our system can significantly improve transfer performance for both weakly supervised and unsupervised tasks. Visualizations demonstrate its strong ability to perceive human positions, outperforming other popular domain adaptation methods. Therefore, it can promote the widespread of WiFi-based pose estimation to new environments without large training and annotation costs.


%





\ifCLASSOPTIONcaptionsoff
  \newpage
\fi



%

\bibliographystyle{IEEEtran}
\bibliography{reference}

\end{document}